\documentclass[a4paper,fleqn]{cas-dc}

\usepackage[authoryear]{natbib}

\usepackage{amsmath,amsfonts}
\usepackage{booktabs}
\usepackage{hyperref}
\usepackage{amsfonts}
\usepackage{xcolor}
\usepackage{color}
\usepackage{amsthm}
\definecolor{mygray}{gray}{.9}
\usepackage{colortbl}

\usepackage{hyperref}

\usepackage{amsmath}
\usepackage{amssymb}

\usepackage{booktabs}
\usepackage{multirow}
\usepackage{makecell}
\usepackage{bbding}

\newcommand{\etal}{\textit{et al.}}

\def\tsc#1{\csdef{#1}{\textsc{\lowercase{#1}}\xspace}}
\tsc{WGM}
\tsc{QE}

\begin{document}
\let\WriteBookmarks\relax
\def\floatpagepagefraction{1}
\def\textpagefraction{.001}

\shorttitle{Multi-scale Contrastive Adaptor Learning for Segmenting Anything in Underperformed Scenes}    
\shortauthors{K. Zhou, Z. Qiu, D. Fu.}  
\title [mode = title]{Multi-scale Contrastive Adaptor Learning for Segmenting Anything in Underperformed Scenes}  



%

\author[1]{Ke Zhou}[]
\credit{Conceptualization, Methodology, Software, Writing - original draft, Writing - review \& editing}

\fnmark[1]





\author[2,3]{Zhongwei Qiu}[]
\fnmark[1]
\credit{Experiment, Experimental analysis, Writing - review \& editing}

\author[1]{Dongmei Fu}[]
\credit{Conceptualization, Supervision, Writing - original draft, Writing - review \& editing}





\affiliation[1]{organization={School of Automation and Electrical Engineering, University of Science and Technology Beijing},
            city={Beijing},
            postcode={100083}, 
            country={China}}

\affiliation[2]{organization={Alibaba DAMO Academy},
            city={Hangzhou},
            postcode={311121}, 
            country={China}}
\affiliation[2]{organization={College of Computer Science and Technology, Zhejiang University},
            city={Hangzhou},
            postcode={311121}, 
            country={China}}


\fntext[1]{Equal contribution}


\begin{abstract}
Foundational vision models, such as the Segment Anything Model (SAM), have achieved significant breakthroughs through extensive pre-training on large-scale visual datasets. Despite their general success, these models may fall short in specialized tasks with limited data, and fine-tuning such large-scale models is often not feasible. Current strategies involve incorporating adaptors into the pre-trained SAM to facilitate downstream task performance with minimal model adjustment. However, these strategies can be hampered by suboptimal learning approaches for the adaptors.
In this paper, we introduce a novel Multi-scale Contrastive Adaptor learning method named MCA-SAM, which enhances adaptor performance through a meticulously designed contrastive learning framework at both token and sample levels. Our Token-level Contrastive adaptor (TC-adaptor) focuses on refining local representations by improving the discriminability of patch tokens, while the Sample-level Contrastive adaptor (SC-adaptor) amplifies global understanding across different samples. Together, these adaptors synergistically enhance feature comparison within and across samples, bolstering the model's representational strength and its ability to adapt to new tasks.
Empirical results demonstrate that MCA-SAM sets new benchmarks, outperforming existing methods in three challenging domains: camouflage object detection, shadow segmentation, and polyp segmentation. Specifically, MCA-SAM exhibits substantial relative performance enhancements, achieving a 20.0\% improvement in MAE on the COD10K dataset, a 6.0\% improvement in MAE on the CAMO dataset, a 15.4\% improvement in BER on the ISTD dataset, and a 7.9\% improvement in mDice on the Kvasir-SEG dataset.
\end{abstract}




\maketitle

\section{Introduction}
Recently, foundational large models have sparked a revolution in the field of artificial intelligence, benefiting from the advanced architecture of the Transformer network, large-scale datasets, and substantial computing resources~\cite{liang2022foundations,zhang2023comprehensive}.
Notably, large language models (LLMs) like BERT~\cite{devlin2018bert}, T5~\cite{raffel2020exploring}, and GPT~\cite{gpt4} have marked a significant breakthrough within the field.
Within the computer vision community, large vision models (LVMs) such as ViT-22B~\cite{dehghani2023scaling}, CLIP~\cite{radford2021learning}, and SAM~\cite{kirillov2023segment} have exhibited tremendous potential.
However, retraining these large models from scratch is impractical for most researchers. Therefore, exploring efficient ways to leverage these foundational large models to assist downstream tasks represents a crucial challenge.

Segmenting Anything Model (SAM)~\cite{kirillov2023segment}, a substantial vision model pre-trained on extensive datasets, is investigated across various downstream vision tasks. Many parameters efficient fine-tuning methods aim to transfer SAM into individual domains by fine-tuning part parameters of SAM~\cite{ma2023segment}, adding trainable adaptors~\cite{chen2023sam,liu2022cada}, or employing prompt tuning~\cite{le2019anabranch}. Adaptor learning, which adds simple adaptors (e.g. two linear layers with an activation layer) into the original SAM network, can adapt the SAM model to different task scenarios and maintain the original strong perception and recognition ability. Thus, it is a popular paradigm for transferring LVMs to downstream tasks.

\begin{figure}[t]
  \centering
  \includegraphics[width=0.9\columnwidth]{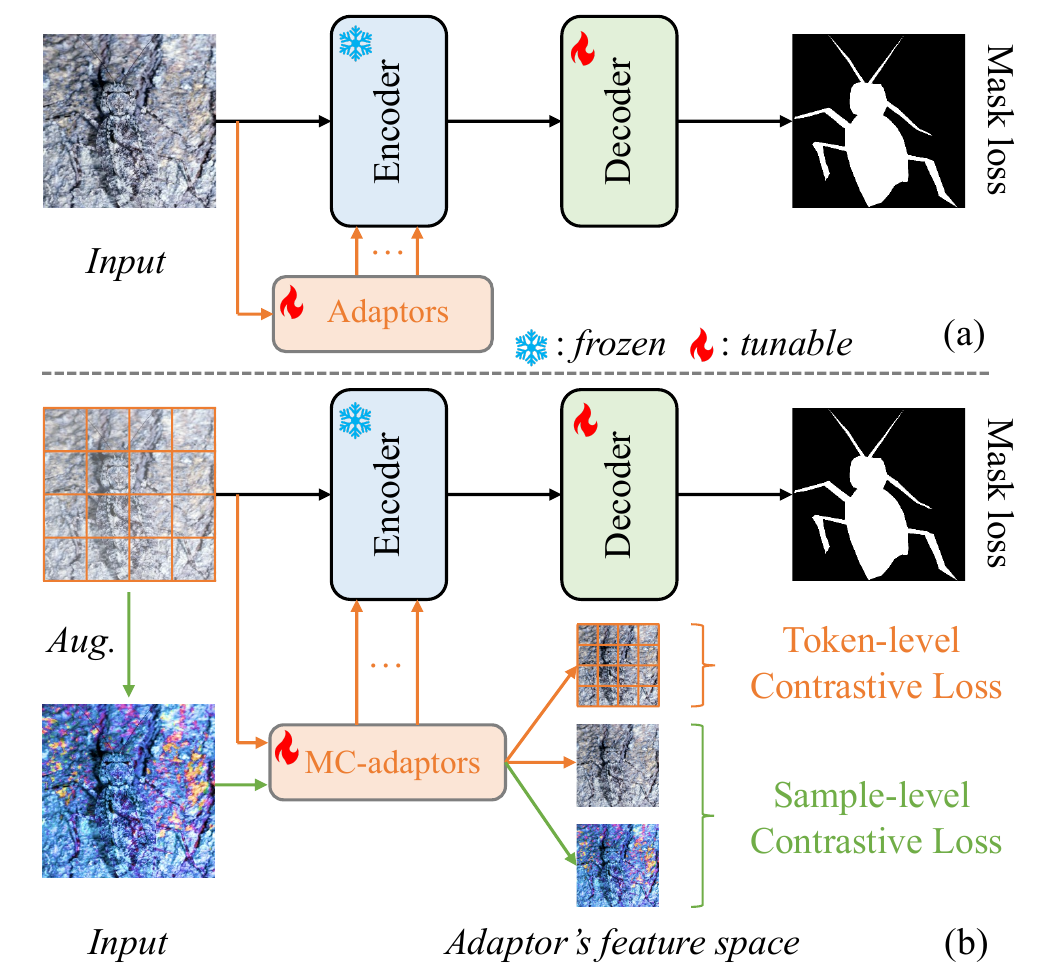}

  \caption{The illustrations of (a) current adaptor learning framework like SAM-adaptor~\cite{chen2023sam} and (b) our multi-scale contrastive adaptor learning framework (MCA-SAM), which consists of encoder, decoder, and Multi-scale Contrastive adaptors (MC-adaptors).  }
  \label{fig:f1}
\end{figure}

The typical framework of current adaptor-based methods like \cite{chen2023sam,liu2023explicit}, is illustrated in Figure \ref{fig:f1} (a). Their pipeline commonly includes a frozen encoder, a tunable decoder, and adaptors. During the training phase, adaptors are integrated into the encoder of the SAM, and a mask loss is employed to fine-tune both the adaptors and the decoder. Although these methods outperform the original SAM in atypical scenarios, their effectiveness is inherently limited by the availability of data in downstream tasks. The limitation stems from an existing adaptor learning strategy that, while parameter-efficient, does not maximize learning efficiency. This paper addresses this shortfall by concentrating on the study of more effective adaptor learning methods, to optimize both the performance and adaptability of SAM across a wider range of tasks.

In recent years, to improve the quality of representation learning, two leading methodologies have shown significant advancements: Masked Image Modeling (MIM) \cite{he2022masked,xie2022simmim} and Contrastive Learning (CL) \cite{oord2018representation,bachman2019learning}. MIM enhances holistic representations by masking patches of the input image and subsequently reconstructing them, whereas CL operates in the feature space to decrease the distance between representations of samples from the same image and increase it for those from different images.
Nonetheless, there is a scarcity of research focused on enhancing the quality of adaptor learning for the transfer of the SAM. While Zhang et al. \cite{zhang2022contrastive} have improved the robustness of zero-shot classification for Large Vision Models (LVMs) by incorporating a contrastive learning loss among the sample embeddings generated by the adaptor, the development of strategies that rapidly advance the representational learning capacities of adaptors remains largely uncharted. Furthermore, existing work~\cite{zhang2022contrastive} tends to concentrate on sample-level contrast, thereby overlooking token-level contrast, which could offer more granular improvements for the contrastive learning capabilities of adaptors.

To address the aforementioned challenges, we introduce a novel Multi-scale Contrastive Adaptor learning framework to enhance the adaptability of the SAM, denoted as MCA-SAM. As illustrated in Figure \ref{fig:f1} (b), MCA-SAM differs from the SAM-adaptor approach \cite{chen2023sam} by employing Multi-scale Contrastive adaptors (MC-adaptors) that bolster SAM's discriminative ability and adaptability at both the token and sample levels. The MC-adaptor consists of Token-level Contrastive adaptors (TC-adaptor) and Sample-level Contrastive adaptors (SC-adaptor). During training, Both local token-level and global sample-level contrastive losses are applied on the features generated by the adaptors. This dual application of contrastive losses fosters more robust and effective representational learning, significantly improving the adaptation ability of SAM in challenging downstream scenarios.

Our contributions can be summarized as follows:
\begin{itemize}
    \item We propose MCA-SAM, a novel representation learning framework that integrates adaptor and contrastive learning for segmenting anything model, effectively enhancing SAM's transferability to underperforming scenarios.
    \item  We propose a multi-scale contrastive adaptor that incorporates token-level and sample-level adaptor contrastive learning, aimed at enhancing the perceptual acuity and discriminative ability among local image patches and distinct samples.
    \item Comprehensive experiments across four benchmarks in three challenging scenarios demonstrate that MCA-SAM surpasses prior methods by significant margins.
\end{itemize}

\section{Related Work}
In this section, we begin by reviewing the literature on foundational vision models. Subsequently, we delve into the research concerning the adaptation of large vision models and the application of contrastive learning techniques to this process. Finally, we will present an overview of three specific scenarios in which large vision model (SAM) typically underperforms: camouflage object detection, shadow segmentation, and polyp segmentation.

\subsection{Foundational Vision Models}
Recently, large-scale foundational models~\cite{radford2021learning,gpt4,ma2023segment} have demonstrated significant advancements within the AI community, especially in the natural language processing (NLP) community and computer vision (CV) community.
Large language models such as BERT~\cite{devlin2018bert}, T5~\cite{raffel2020exploring}, and GPT~\cite{gpt4} employ masked language modeling or generative pre-training techniques to train substantial Transformer-based architectures on extensive datasets. These models are imbued with the capacity for zero-shot transfer learning across a multitude of downstream tasks.

Drawing inspiration from the success in the NLP field, the pre-training of large-scale foundational models in the CV domain has also flourished.
Similar to the works in NLP, many researches like ViT-G~\cite{zhai2022scaling}, ViT-22B~\cite{dehghani2023scaling}, Swin Transformer V2~\cite{liu2022swin}, and VideoMAE V2~\cite{wang2023videomae} try to enlarge the scale size of model, which could improve the capacity of model on handling complex downstream tasks~\cite{Qiu_2023_CVPR}.
On the other hand, some prior works like MAE~\cite{he2022masked}, CAE~\cite{chen2023context}, SimMIM~\cite{xie2022simmim}, BEiT V2~\cite{peng2022beit}, EVA~\cite{fang2023eva} explore the different self-supervised pre-training strategies for large vision models.

To tackle the multimodal data, some notable works like CLIP~\cite{radford2021learning} and ALIGN~\cite{jia2021scaling} align the vision and language representations in a feature domain by contrastive learning~\cite{oord2018representation}. These efforts have laid the groundwork for research on cross-modal technologies.

Just as GPT~\cite{gpt4} has brought tremendous surprises to the NLP field, the computer vision community still awaits the emergence of groundbreaking foundational and universal large models. 
Benefiting from significant investments in datasets and computational resources, the Segmenting Anything Model (SAM)~\cite{kirillov2023segment} has brought notable progress to the field of computer vision.
SAM pre-trains a suite of large-scale ViT-based models on an expansive vision dataset comprising over 11 million images with more than 1 billion masks, thereby enabling zero-shot segmentation and demonstrating substantial potential in the field of computer vision.
However, these pre-trained models often exhibit subpar performance in certain atypical scenarios, such as detecting camouflaged objects, segmenting shadows, and segmenting polyps.

\subsection{Adaptation of Large Vision Models}
With the increasing model size, adapting large models into downstream tasks is first used in the NLP field~\cite{houlsby2019parameter,sung2021training,zaken2022bitfit,qing2023mar}. They aim to reduce the computational costs by fine-tuning partial parameters of large language models and hope to reach comparable performances of the full fine-tuning strategy.
Recently, these parameter-efficient fine-tuning methods~\cite{bahng2022visual,jie2022convolutional,gao2022visual,chen2023sam,liu2023explicit,peng2024sam,Peng_2024_CVPR} have also been utilized in computer vision.
Particularly, EVP~\cite{liu2023explicit} and ViT-Adapter~\cite{chen2022vision} insert adaptors into the ViT model to transfer it into downstream tasks.
Moreover, MedSAM~\cite{ma2023segment} and SAM-adaptor~\cite{chen2023sam} adapt the pre-train SAM to downstream scenarios by fine-tuning the whole SAM on medical datasets or adding adaptors to SAM.
Although adaptor-based methods are parameters-efficient, they still rely on the data size of downstream tasks to improve the discriminability of SAM and their learning strategy can be improved.
In this paper, we investigate the transfer learning capabilities of SAM and propose an innovative adaptor-based learning approach to enhance the adaptability of SAM.

\subsection{Contrastive Learning}
Contrastive learning~\cite{ki2021contrastive} is a self-supervised learning approach that aims to learn a discriminative representation of an image by pulling together the representations of positively paired images while pushing away those of negatively paired images.
Existing works like SimCLR~\cite{chen2020simple}, MoCo~\cite{he2020momentum}, BYOL~\cite{grill2020bootstrap}, and DINO~\cite{caron2021emerging} build various positive image pairs and negative samples by different augmentation strategies to pre-train large vision Transformer. Some pixel cross-image pixel contrast methods~\cite{wang2021exploring,zhou2022rethinking} are proposed to enhance semantic segmentation.
Their successes show the huge potential of contrastive learning, while few works explore leveraging contrastive learning on adaptor learning. Although Zhang \etal~\cite{zhang2022contrastive} add contrastive loss on the sampling embeddings to improve the zero-shot classification robustness of the vision Transformer, it remains the challenge how to use contrastive learning to improve the adaptation of LVMs. In this work, we aim to tackle this problem by proposing a multi-scale contrastive adaptor learning approach.

\begin{figure*}[t]
  \centering
  \includegraphics[width=\linewidth]{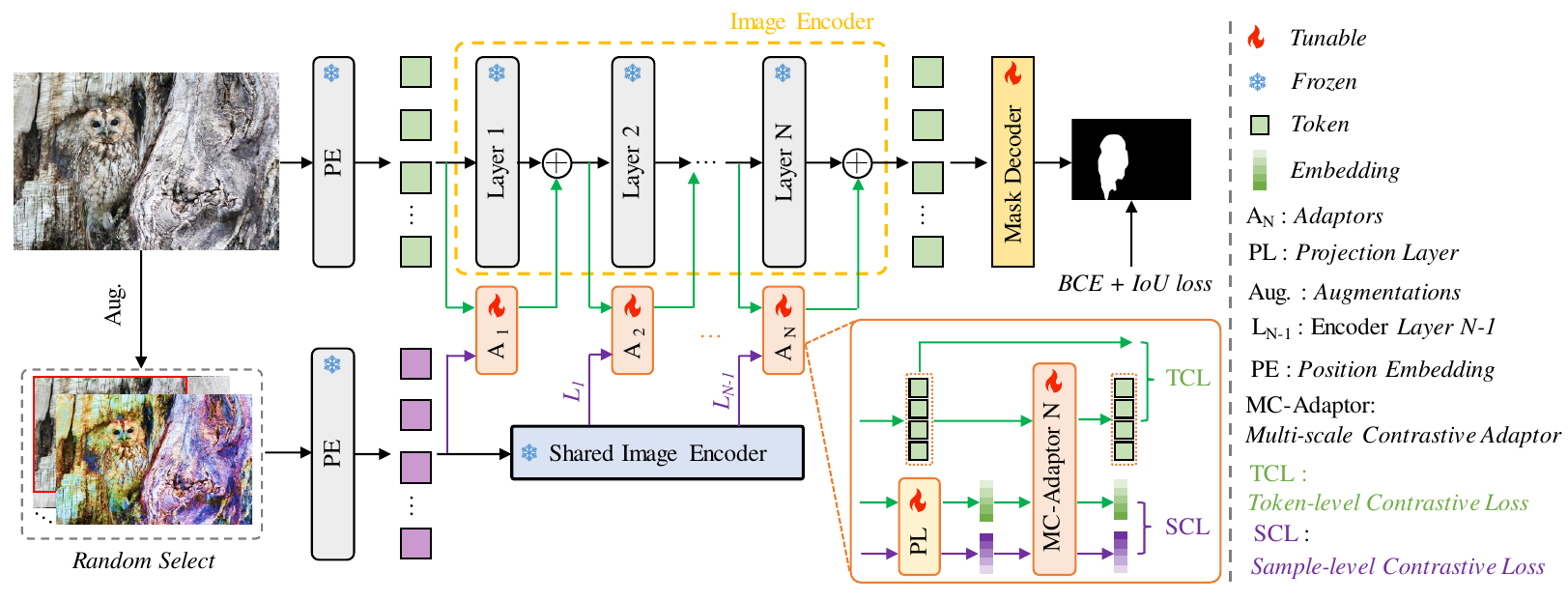}

  \caption{The framework of Multi-scale Contrastive Adaptor learning for SAM (MCA-SAM). During training, the MC-adaptors are inserted into each Transformer layer of SAM. The parameters of the image encoder in SAM are frozen, while only the parameters of adaptors and mask decoder are tunable.
  MC-adaptors include contrastive adaptors in both the token level and sample level, with the supervision of token-level contrastive loss and sample-level contrastive loss. During inference, the inputs of each Transformer layer are the summation of the outputs of the last layer and the outputs of the current adaptor. 
  $\otimes$ represents element-wise sum.}
  \label{fig:framework}
\end{figure*}

\subsection{Typical Underperformed Scenes for SAM}
Although SAM~\cite{kirillov2023segment} shows a strong capacity for zero-shot transferring on many downstream tasks~\cite{cheng2023segment,10239462,9878466}, it underperforms in some scenarios, such as camouflage object detection, shadow segmentation, and polyp segmentation.
Camouflage object detection~\cite{le2019anabranch,mei2021camouflaged,fan2020camouflaged,zhai2022deep,li2023camouflaged} requires distinguishing the foreground and background in an image, which is a challenging task since foreground and background have similar visual patterns.
Shadows are typically caused by factors such as insufficient exposure and object obstruction~\cite{wang2019moving,niu2022boundary,liu2023explicit,ahn2023domain}, making it a challenging task.
Polyp Segmentation~\cite{jha2020kvasir,jain2023coinnet,su2023accurate} is a typical medical image segmentation task, which is difficult for zero-shot transferring since the target and background have similar visual features.
The three tasks mentioned above serve as challenging benchmarks, where existing large vision models tend to consistently underperform~\cite{ma2023segment,chen2023sam}.
Thus, many works like MedSAM~\cite{ma2023segment}, SAM-adaptor~\cite{chen2023sam}, and EVP~\cite{liu2023explicit} propose different transferring methods for SAM in these tasks. In this work, we propose an efficient contrastive adaptor learning method to transfer SAM into downstream tasks and evaluate our approach in the three scenarios.

\section{Method}
In this section, we will first introduce the framework of the proposed method in Section \ref{sec:framework}, which consists of SAM and multi-scale contrastive adaptors.
Then, the main components of multi-scale contrastive adaptor are introduced in Section \ref{sec:tca} and Section \ref{sec:sca}, respectively.
Finally, the loss functions for fine-tuning adaptors are shown in Section \ref{sec:loss}.

\subsection{Framework}
\label{sec:framework}
The framework of Multi-scale Contrastive Adaptor learning for SAM (MCA-SAM) is illustrated in Figure \ref{fig:framework}.
Given an image $I$, the image tokens with position embedding are sent into the image encoder of SAM to learn the representations of the image. Then, the target mask is decoded by the mask decoder of SAM.
The image encoder consists of several Transformer layers according to different model sizes (e.g. ViT-B, ViT-L, and ViT-H). Assuming there are $N$ Transformer layers in the image encoder, the forward process of SAM can be formulated as:
\begin{equation}
    M = D(E(T(I) + PE(I))),
\end{equation}
where $M$, $E$, $D$, $T(\cdot)$, and $PE(\cdot)$ are the output mask, image encoder, mask decoder, tokenization operation, and position embedding layer, respectively.
Image encoder $E$ includes $N$ standard vision Transformer layer $L$~\cite{dosovitskiy2020image}, which can be written as:
\begin{equation}
\label{eq:transformer_layer}
    X^i = L^i(X^{i-1}), i \in [1,N],
\end{equation}
where $X^0 = T(I) + PE(I)$ and $X$ represents the visual tokens.

For adaptor learning, the adaptors are inserted into each layer of the image encoder as previous works~\cite{chen2023sam,liu2023explicit}. The process of Equation \ref{eq:transformer_layer} can be recapped as:
\begin{equation}
    X^i = L^i(X^{i-1}) +  A^{i}(X^{i-1}), i \in [1,N],
\end{equation}
where $A$ represents the proposed Multi-scale Contrastive Adaptors (MC-Adaptors), which include Token-level Contrastive Adaptors (TC-Adaptors, denoted as $A_{TC}$), and Sample-level Contrastive Adaptors (SC-Adaptors, denoted as $A_{SC}$) that are introduced in the following.

\subsection{Token-level Contrastive Adaptor Learning}
\label{sec:tca}
\begin{figure}[t]
  \centering
  \includegraphics[width=0.95\linewidth]{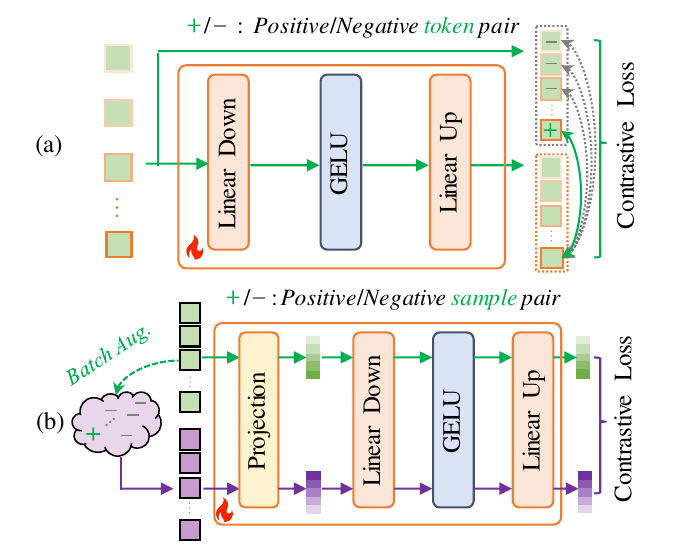}

  \caption{The architecture of token-level contrastive adaptor (TC-adaptor) and sample-level contrastive adaptor (SC-adaptor). (a) TC-adaptor enhances the discriminability of SAM among local spatial tokens. (b) SC-adaptor enhances the discriminability of SAM among batch samples.}
  \label{fig:tca_sca}
\end{figure}

Token-level Contrastive Adaptor (TC-adaptor) is designed to enhance the discriminability of adaptor among spatial tokens. The architecture of TC-adaptor is shown in Figure \ref{fig:tca_sca} (a). TC-adaptor $A_{TC}$ including two linear layers and one activation layer, which can be written as:
\begin{equation}
\label{eq:tc}
    Y^i = A^i_{TC}(X^i) = f_{up}(\zeta(f_{down}(X^i))), i\in [1,N],
\end{equation}
where $Y$ represents the output of TC-adaptor $A_{TC}$. $f_{up}$, $f_{down}$, and $\zeta$ are a linear layer of reducing feature dimension, a linear layer of increasing feature dimension, and the GELU~\cite{hendrycks2016gaussian} activation layer, respectively.

For token-level contrastive learning, the output tokens $Y$ will be applied to compute the contrastive loss with the input tokens $X$ of the TC-adaptor.
The objective of token-level contrastive learning is to minimize the distance between identical tokens before and after undergoing transformation by the TC adapter, while concurrently maximizing the distance between distinct tokens that have undergone the same TC-adaptor transformation.
Thus, the training objective of the TC-adaptor can be formulated as:
\begin{equation}
   CL_t(X, Y) = -log \frac{exp(\rho^+_t / \tau )}{exp(\rho^+_t / \tau) + \sum_{j=1}^{K-1}exp(\rho^-_t / \tau) },
\end{equation}
where $CL_t$, $\rho^+_t$, $\rho^-_t$, $\tau$ represent the contrastive loss for TC-adaptor, cosine similarity of positive token-level pairs, cosine similarity of negative token-level pairs, and a temperature constant, respectively. $K$ is the token number.
The cosine similarity for token-level pairs is computed as:
\begin{equation}
    \rho^{j\tilde{j}} = \frac{x^j\cdot y^{\tilde{j}}}{||x^j||\cdot ||y^{\tilde{j}}||}, j,\tilde{j} \in [1,K],x^j \in X, y^{\tilde{j}} \in Y,
\end{equation}
where $\rho^{j\tilde{j}}$ represents the cosine similarity between $j$-th token from $X$ and the $\tilde{j}$-th token from $Y$.

During training, $CL_t$ is used as an optimization loss to train the TC-adaptor. 

\subsection{Sample-level Contrastive Adaptor Learning}
\label{sec:sca}
Sample-level Contrastive Adaptor (SC-adaptor) is designed to strengthen the global discriminability of adaptors among different samples in a batch.
The architecture of the SC-adaptor is shown in Figure \ref{fig:tca_sca} (b). SC-adaptor consists of a projection layer, two linear layers, and an activation layer.
Given a batch of images $\text{I} = \{I^b, b\in [1, B]\}$, where $B$ is the batch size, they are first transferred into another view by data augmentations like ColorJitter, Grayscale, and RandomShift. The augmented images are denoted as $\text{I}_{aug} = \{\Psi({I}^b), b \in [1,B]\}$, where $\Psi$ is a random selected augmentation operation.
Similar to TC-adaptor, image tokens $X$ and $X_{aug}$ are extracted from $\text{I}$ and $\text{I}_{aug}$, respectively.
Then, $X$ and $X_{aug}$ are sent into the SC-adaptor to compute the contrastive loss at a global sample level.

For the sample-level contrastive learning, image tokens $X$ and $X_{aug}$ are projected into embeddings by projection layer $f_{p}$, respectively. They are further sent into the adaptor and output features $\epsilon$ and $\epsilon_{aug}$. This process can be formulated as:
\begin{equation}
\label{eq:sc}
    \epsilon^b = A_{SC}^b = f_{up}(\zeta(f_{down}(f_p(X^b)))), b \in [1, B],
\end{equation}
where $A_{SC}$ represents the SC-adaptor that is similar to Equation \ref{eq:tc}. However, the SC-adaptor is applied at the sample level while not the token level.
The augmented embeddings $\epsilon_{aug}$ can obtained following Equation \ref{eq:sc}.

For an image $I^b$ in a batch, the corresponding augmented image $I^b_{aug}$ is its positive sample pair while others are negative pairs. 
Thus, the training objective of the SC-adaptor can be formulated as:
\begin{equation}
    CL_{s}(\epsilon, \epsilon_{aug})= -log \frac{exp(\rho^+_s / \tau )}{exp(\rho^+_s / \tau) + \sum_{b=1}^{B-1}exp(\rho^-_s / \tau) },
\end{equation}
where  $CL_s$, $\rho^+_s$, $\rho^-_s$, $\tau$ represent the contrastive loss for SC-adaptor, cosine similarity of positive sample-level pairs, cosine similarity of negative sample-level pairs, and a temperature constant, respectively.
The cosine similarity for token-level pairs is computed as:
\begin{equation}
    \rho^{b\tilde{b}} = \frac{\epsilon^b\cdot \epsilon_{aug}^{\tilde{b}}}{||\epsilon^b||\cdot ||\epsilon_{aug}^{\tilde{b}}||}, b,\tilde{b} \in [1,B],\epsilon^b \in \epsilon, \epsilon_{aug}^{\tilde{b}} \in \epsilon_{aug},
\end{equation}
where $\rho^{b\tilde{b}}$ represents the cosine similarity between $b$-th sample from $\epsilon$ and the $\tilde{b}$-th sample from $\epsilon_{aug}$.

During training, $CL_s$ is used the an optimization loss to train SC-adaptor.

\subsection{Loss Function}
\label{sec:loss}
During training, images $\text{I}$ are sent into the image encoder to encode the image tokens and to further decode mask by mask decoder.
Meanwhile, the images $\text{I}$ also are transferred into the augmentation domain to extract the augmented image tokens by a shared image encoder.
The multi-scale contrastive adaptors are inserted into each layer of the image encoder.
The token-level contrastive loss and sample-level contrastive loss are applied to the multi-scale contrastive adaptors.
Following previous works~\cite{chen2023sam,liu2023explicit}, the BCE loss and IoU loss are applied to the mask decoder.
The parameters of the Transformer layer in the image encoder are frozen.
Therefore, the total loss for fine-tuning the network can be written as:
\begin{equation}
    \mathcal{L} = \mathcal{L}_{BCE}(M, M_{gt}) + \mathcal{L}_{IoU}(M, M_{gt}) + CL_t + CL_s,
\end{equation}
where $\mathcal{L}$, $\mathcal{L}_{BCE}$, and $\mathcal{L}_{IoU}$ represent the total loss, BCE loss, and IoU loss, respectively. $M$ and $M_{gt}$ are the predicted mask and ground-truth mask. $CL_t$ and $CL_s$ are the corresponding contrastive loss for TC-adaptors and SC-adaptors in Equation \ref{eq:tc} and Equation \ref{eq:sc}, respectively.

During inference, the fine-tuned adaptors are inserted into each layer of the image encoder, which means that there are $N$ adaptors for an image encoder with $N$ Transformer layer. The image tokens are inputted into adaptors and the output tokens are used to add to the output tokens of the current Transformer layer. The augmentations are not used during inference.

\section{Experiments}
In this section, we introduce the implementation details and experimental results. Specifically, the evaluation benchmarks and implementation details are introduced in Section \ref{sec:exp_data} and Section \ref{sec:exp_im_detail}. Section \ref{sec:exp_main_res} presents the main results and comparison. Visualization comparison is shown in Section \ref{sec:exp_vis}. The ablation study and analysis of MCA-SAM are introduced in Section \ref{sec:exp_abalation}. Finally, the failure cases are discussed in Section \ref{sec:fail}.

\subsection{Datasets and Evaluation Metrics}
\label{sec:exp_data}
Following previous works~\cite{liu2023explicit,chen2023sam}, four datasets in three underperformed scenes are used to evaluate the performance of MCA-SAM. Camouflage object detection, shadow segmentation, and polyp segmentation represent three typical scenarios where SAM often underperforms, which are used in this paper for a fair comparison with prior methods. The widely used datasets include the COD10K~\cite{fan2020camouflaged} and CAMO~\cite{le2019anabranch} datasets for camouflage object detection, the ISTD~\cite{wang2018stacked} dataset for shadow segmentation, and the Kvasir-SEG~\cite{jha2020kvasir} dataset for polyp segmentation.

\subsubsection{COD10K dataset for Camouflage Object Detection}
COD10K~\cite{fan2020camouflaged} is a widely-used dataset curated for camouflage object detection, partitioned into training and testing subsets, comprising 3,040 and 2,026 images respectively. Following previous works~\cite{liu2023explicit,chen2023sam}, Structure measure (S-measure $S_\alpha$), mean Enhanced-alignment measure (E-measure $E_\phi$), and Mean Absolute Error (MAE) are used for evaluation. MAE measures the pixel error as:
\begin{equation}
    MAE = \frac{1}{W\times H}\sum^W_{i=1}\sum^H_{j=1}|G(i,j)-S(i,j)|,
\end{equation}
where $S \in [0,1]^{W\times H}$ and $G \in \{0,1\}^{W\times H}$ represent normalized prediction map and binary ground-truth mask. $H$ and $W$ are the height and width of the mask.

$S_\alpha$ measures the structural similarity between $S$ and $G$, which is defined as:
\begin{equation}
    S_\alpha = \alpha \times S_o + (1-\alpha) \times S_r,
\end{equation}
where $S_o$ and $S_r$ are object-aware and region-aware similarity, respectively, as defined in \cite{fan2017structure}. $\alpha$ is set to 0.5.

$E_\phi$ is defined as:
\begin{equation}
    E_\phi = \frac{1}{W\times H} \sum^W_{i=1}\sum^H_{j=1}\phi_S(i,j),
\end{equation}
where $\phi_S$ represents the enhanced alignment matrix, which computes the correlation between $S$ and $G$ as \cite{fan2018enhanced}.

\subsubsection{CAMO dataset for Camouflage Object Detection}
CAMO~\cite{le2019anabranch} is another widely-used dataset for camouflage object detection, which includes 1,250 training images and 250 testing images.
Similar to the COD10K dataset and following \cite{liu2023explicit,chen2023sam}, Structure measure (S-measure $S_\alpha$), mean Enhanced-alignment measure (E-measure $E_\phi$), and Mean Absolute Error (MAE) are used for evaluation in this dataset. 

\subsubsection{ISTD dataset for Shadow Segmentation}
ISTD~\cite{wang2018stacked} is a commonly used dataset for shadow segmentation, which contains a training set of 1,330 images and a testing set of 540 images.
For a fair comparison with previous methods~\cite{wang2018stacked,liu2023explicit,chen2023sam},  balance error rate (BER) is used for the evaluation of shadow segmentation. The BER is defined as:
\begin{equation}
    BER = 1 - \frac{1}{2}(\frac{TP}{TP+FN} + \frac{TN}{TN+FP}),
\end{equation}
where TP, TN, FP, and FN represent the pixel-level true positive, true negative, false positive, and false negative, respectively.

\subsubsection{Kvasir-SEG dataset for Polyp Segmentation}
Polyp segmentation is a medical image segmentation task. Kvasir-SEG dataset~\cite{jha2020kvasir} is a representative dataset and contains 1,000 images for training and 196 images for testing.
Following the widely used and officially suggested metrics for medical image segmentation, the mean Dice coefficient (mDice) and mean Intersection over Union (mIoU) are used for evaluation in this dataset.
Dice is computed as:
\begin{equation}
    Dice = \frac{2 \times |S \cap G|}{|S| + |G|}.
\end{equation}

IoU is computed as:
\begin{equation}
    IoU = \frac{|S \cap G|}{|S \cup G|}.
\end{equation}

In experiments, the above four datasets are used to evaluate the performance of MCA-SAM. All ablation studies are conducted on the Kvasir-SEG dataset~\cite{jha2020kvasir}.

\subsection{Implementation Details}
\label{sec:exp_im_detail}
\paragraph{\textbf{Network.}}
In the experiments, we use the Segmenting Anything Model (SAM)~\cite{kirillov2023segment} as the target foundational vision model, including three types of backbone networks: ViT-B, ViT-L, and ViT-H~\cite{dosovitskiy2020image}. Following prior adaptor-based methods \cite{liu2023explicit,chen2023sam}, all the used ViT backbone networks are pre-trained by SAM \cite{kirillov2023segment}.
For the sample-level and token-level adaptors, each adaptor consists of 1 downsample Linear layer, 1 GELU activation layer, and 1 upsample Linear layer. For sample-level adaptors, an extra Linear layer is added to project tokens into sample representation. The number of adaptors is 12, 24, 32 for ViT-B, ViT-L, and ViT-H, respectively.

\paragraph{\textbf{Training.}}
In the experiments, the used optimizer is AdamW and the initial learning rate is 2e-4. The learning rate decreases to 1e-7 by the Cosine decay schedule. The total training epochs are 20, 90, and 120 for camouflaged object detection, shadow segmentation, and polyp segmentation, respectively. PyTorch is used to implement the experiments and 4 Nvidia A800 GPUs are used to train MCA-SAM. The temperature constant for contrastive loss is set to 0.1. The scale factor to reduce and increase the dimension of embedding in $f_{down}$ and $f_{up}$ is 32, thus the embedding dimensions of adaptors are variational for different ViT sizes. For a fair comparison with the previous method~\cite{liu2023explicit,chen2023sam}, the image size is $1024\times 1024$. During training, the loss weights for BCE, IoU, and contrastive loss are 1.

\subsection{Main Quantitative Results}
\label{sec:exp_main_res}
To verify the advancements of MCA-SAM, experiments are carried out on four distinct datasets, and comparisons are made with existing state-of-the-art methods.

\subsubsection{Camouflage Object Detection on COD10K dataset}
Camouflage object detection is a challenging task since the target camouflage objects show similar visual features to the background, which requires strong discriminability for models to achieve good results. MCA-SAM is fine-tuned on this dataset to inspect its capacity. 
The results of MCA-SAM based on ViT-B, ViT-L, and ViT-H are shown in Table \ref{tab:cod}. The three types of MCA-SAM achieve 0.103, 0.073, and 0.066 in MAE on the test set of COD10K~\cite{fan2020camouflaged}. MCA-SAM (ViT-H) outperforms the previous SOTA method (SAM-adaptor~\cite{chen2023sam}) by 0.020 in MAE, with relative improvements of 20\%.
For structure metrics, MCA-SAM also outperforms existing approaches and reaches $S_\alpha$ of 0.917 and $E_\phi$ of 0.930.
These results demonstrate the strong ability of MCA-SAM to distinguish camouflage objects.

\begin{table}
  \caption{Quantitative Result for Camouflage Object Detection on COD10K~\cite{fan2020camouflaged} dataset.}
  \centering
  \renewcommand\tabcolsep{6pt}
  \resizebox{\columnwidth}{!}{
  \begin{tabular}{c|c|c|c}
    \toprule
    Method & $S_\alpha \uparrow$ & $E_\phi \uparrow$& MAE $\downarrow$\\
    \midrule
    SINet~\cite{fan2020camouflaged} & 0.771 & 0.806 & 0.051 \\
    RankNet~\cite{lv2021simultaneously} & 0.767 & 0.861 & 0.045 \\
    JCOD~\cite{li2021uncertainty} & 0.800 & 0.872 & 0.041 \\
    PFNet~\cite{mei2021camouflaged} & 0.800 & 0.868 & 0.040 \\
    FBNet~\cite{lin2023frequency} & 0.809 & 0.889 & 0.035\\
    \midrule
    SAM (ViT-H)~\cite{kirillov2023segment} & 0.783 & 0.798 & 0.050 \\
    EVP (ViT-H)~\cite{liu2023explicit} & 0.846 & 0.895 & 0.059 \\
    SAM-adaptor (ViT-H)~\cite{chen2023sam} & 0.883 & 0.918 & 0.025\\
    \midrule
    MCA-SAM (ViT-B) (Ours)& 0.885 & 0.899 & 0.038\\
    MCA-SAM (ViT-L) (Ours)& 0.910 & 0.923 & 0.029\\
    MCA-SAM (ViT-H) (Ours)& \textbf{0.917} & \textbf{0.930} & \textbf{0.020}\\
        
    \bottomrule
  \end{tabular}
  }
  \label{tab:cod}
\end{table}

\subsubsection{Camouflage Object Detection on CAMO dataset}
CAMO~\cite{le2019anabranch} dataset serves as another widely adopted benchmark for assessing the discriminative capabilities of various methods in the context of camouflage object detection.
Similar to the experiments conducted on the COD10K dataset, we fine-tuned MCA-SAM on this dataset, and the results are presented in Table \ref{tab:camo}.
MCA-SAM, utilizing the ViT-H backbone network, achieves 0.851, 0.910, and 0.023 in $S_\alpha$, $E_\phi$, and MAE, respectively.
It surpasses the SAM-adaptor~\cite{chen2023sam} and leads a new SOTA results on this dataset, with a relative improvement of 6\% in MAE.
These results once again underscore the robust performance of MCA-SAM in camouflage object detection. The leading outcomes on both datasets further demonstrate the strong generalization capability of the MCA-SAM approach.

\begin{table}
  \caption{Quantitative Result for Camouflage Object Detection on CAMO~\cite{le2019anabranch} dataset.}
  \centering
  \renewcommand\tabcolsep{6pt}
  \resizebox{\columnwidth}{!}{
  \begin{tabular}{c|c|c|c}
    \toprule
    Method & $S_\alpha \uparrow$ & $E_\phi \uparrow$ & MAE $\downarrow$\\
    \midrule
    SINet~\cite{fan2020camouflaged} & 0.751 & 0.771 & 0.100 \\
    RankNet~\cite{lv2021simultaneously} & 0.712 & 0.791 & 0.104 \\
    JCOD~\cite{li2021uncertainty} & 0.792 & 0.839 & 0.082 \\
    PFNet~\cite{mei2021camouflaged} & 0.782 & 0.852 & 0.085\\
    FBNet~\cite{lin2023frequency} & 0.783 & 0.839 & 0.081\\
    \midrule
    SAM (ViT-H)~\cite{kirillov2023segment} & 0.684 & 0.687 & 0.132\\
    EVP (ViT-H)~\cite{liu2023explicit} & 0.843 & 0.907 & 0.029 \\
    SAM-adaptor (ViT-H)~\cite{chen2023sam} & 0.847 & 0.873 & 0.070\\
    \midrule
    MCA-SAM (ViT-B) (Ours)& 0.772 & 0.798 & 0.103 \\
    MCA-SAM (ViT-L) (Ours)& 0.834 & 0.864 & 0.073 \\
    MCA-SAM (ViT-H) (Ours)& \textbf{0.851} & \textbf{0.910} & \textbf{0.023} \\

    \bottomrule
  \end{tabular}
  }
  \label{tab:camo}
\end{table}

\begin{table}
  \caption{Quantitative Result for Shadow Segmentation on ISTD~\cite{wang2018stacked} dataset.}
  \centering
  \renewcommand\tabcolsep{20pt}
  \resizebox{\columnwidth}{!}{
  \begin{tabular}{c|c}
    \toprule
    Method & BER $\downarrow$ \\
    \midrule
    Stacked CNN~\cite{vicente2016large} & 8.60\\
    BDRAR~\cite{zhu2018bidirectional} & 2.69\\
    DSC~\cite{hu2018direction} & 3.42\\
    DSD~\cite{zheng2019distraction} & 2.17 \\
    FDRNet~\cite{zhu2021mitigating} & 1.55\\
    \midrule
    SAM (ViT-H)~\cite{kirillov2023segment}& 40.51\\
    EVP (ViT-H)~\cite{liu2023explicit} & 1.35\\
    SAM-adaptor (ViT-H)~\cite{chen2023sam}& 1.43\\
    \midrule
    MCA-SAM (ViT-B) (Ours)& 1.98\\
    MCA-SAM (ViT-L) (Ours)& 1.43\\
    MCA-SAM (ViT-H) (Ours)& \textbf{1.21}\\
    \bottomrule
  \end{tabular}
  }
  \label{tab:istd}
\end{table}

\subsubsection{Shadow Segmentation on ISTD dataset}
Shadow segmentation is an uncommon scenario that causes the original SAM~\cite{kirillov2023segment} to underperform.
We perform comparative experiments between conventional shadow segmentation methods~\cite{zheng2019distraction,zhu2021mitigating}, SAM-based methods~\cite{kirillov2023segment,chen2023sam}, and our MCA-SAM, as shown in Table \ref{tab:istd}.
With a fair comparison, MCA-SAM based on ViT-H achieves a BER of 1.21, which outperforms all prior methods with a relative improvement of 15\%.
Additionally, MCA-SAM based on ViT-L achieves a comparable BER of 1.43 with the SAM-adaptor based on ViT-H. This result underscores the effective multi-scale adaptor learning in MCA-SAM.

\subsubsection{Polyp Segmentation on Kvasir-SEG dataset}
To validate the applicability of MCA-SAM across various scenarios, we conducted experiments comparing its performance in polyp segmentation tasks on the Kvasir-SEG dataset~\cite{jha2020kvasir}, which is another uncommon scenario for SAM. 
The results are shown in Table \ref{tab:Polyp}.
With ViT-H as a backbone network, MCA-SAM achieves 0.917 in mDice and 0.845 in mIoU, which outperforms SAM-adaptor~\cite{chen2023sam} by mDice of 0.067 and mIoU of 0.069. The relative improvements are 7.9\% and 8.9\%, showing the super capacity of cross-scale adaptor learning.

\begin{table}
  \caption{Quantitative Result for Polyp Segmentation on Kvasir-SEG~\cite{jha2020kvasir} dataset. $*$ indicates that EVP is trained on Kvasir-SEG by using their codes.}
  \centering
  \renewcommand\tabcolsep{8pt}
  \resizebox{\columnwidth}{!}{
  \begin{tabular}{c|c|c}
    \toprule
    Method  & mDice $\uparrow$ & mIoU $\uparrow$\\
    \midrule
    UNet~\cite{ronneberger2015u} & 0.821 & 0.756 \\
    UNet++~\cite{zhou2018unet++} & 0.824 & 0.753 \\
    SFA~\cite{fang2019selective} & 0.725 & 0.619 \\
    \midrule
    SAM (ViT-H)~\cite{kirillov2023segment}& 0.778 & 0.707 \\
    EVP (ViT-H)$^*$~\cite{liu2023explicit} & 0.849 & 0.773 \\
    SAM-adaptor (ViT-H)~\cite{chen2023sam}& 0.850 & 0.776 \\
    \midrule
    MCA-SAM (ViT-B) (Ours)& 0.836 & 0.748 \\
    MCA-SAM (ViT-L) (Ours)& 0.889 & 0.800 \\
    MCA-SAM (ViT-H) (Ours)& \textbf{0.917} & \textbf{0.845} \\
    \bottomrule
  \end{tabular}
  }
  \label{tab:Polyp}
\end{table}

Quantitative comparative experiments on four datasets consistently demonstrate that MCA-SAM surpasses existing methods, confirming the model's superior performance and validating the effectiveness of the proposed adaptor-contrastive learning mechanism.

\begin{figure*}[t]
  \centering
  \includegraphics[width=\linewidth]{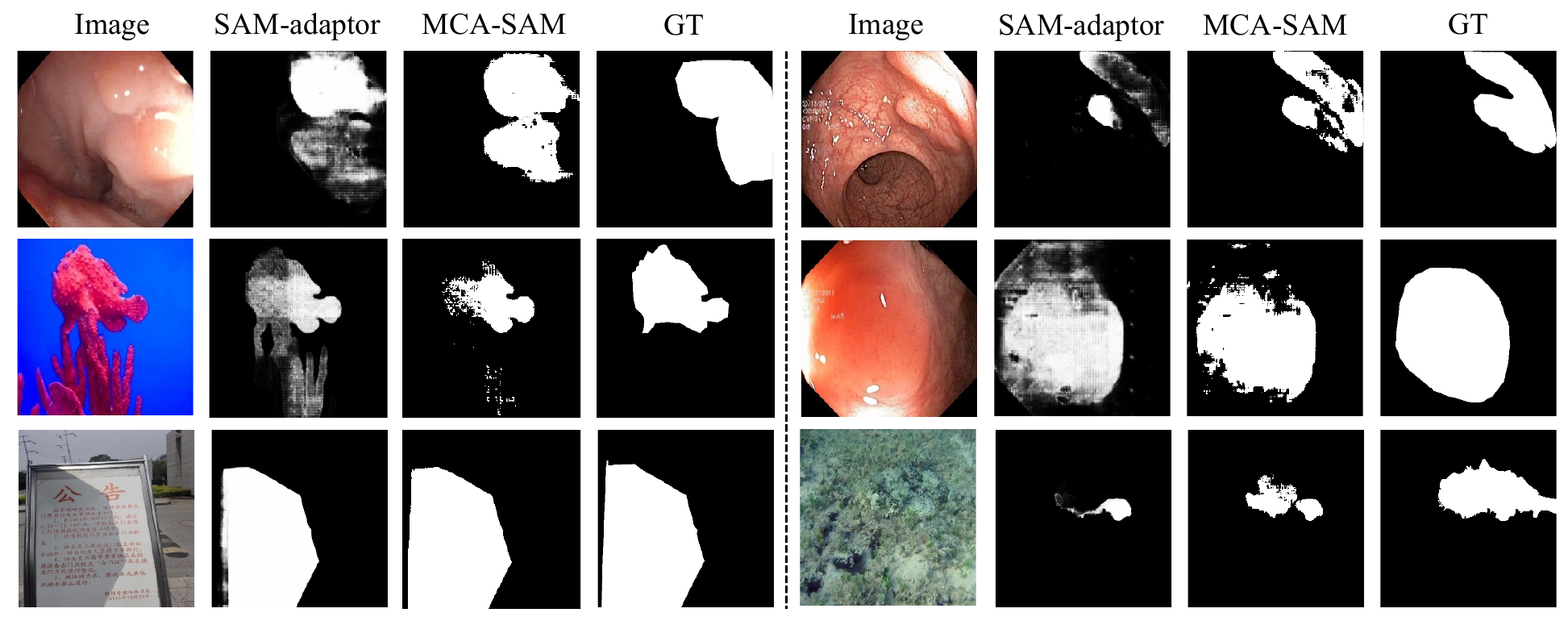}

  \caption{The visualization comparison with SAM-adaptor~\cite{chen2023sam} on the \textbf{extremely challenging cases} from COD10K~\cite{fan2020camouflaged}, CAMO~\cite{le2019anabranch}, ISTD~\cite{wang2018stacked}, and Kvasir~\cite{jha2020kvasir} datasets, respectively. MCA-SAM can localize some pixels that are difficult to distinguish due to the stronger capacity of the multi-scale contrastive adaptors.}
  \label{fig:vis}
\end{figure*}

\subsection{Visulization Comparison}
\label{sec:exp_vis}
To evaluate the effectiveness of MCA-SAM and compare it with prior SAM-adaptor~\cite{chen2023sam}, we visualize the results of MCA-SAM and SAM-adaptor on the extremely challenging cases from four datasets: COD10K~\cite{fan2020camouflaged}, CAMO~\cite{le2019anabranch}, ISTD~\cite{wang2018stacked}, and Kvasir~\cite{jha2020kvasir} datasets.
As shown in Figure \ref{fig:vis}, MCA-SAM can achieve better results than SAM-adaptor in these challenging cases.
These visualization results demonstrate the stronger capacity of multi-scale contrastive adaptors to distinguish these hard samples, which also provides a stronger adaptation ability for SAM to handle these cases.

\begin{table}
  \caption{The comparison of parameters and FLOPs with adaptor-based methods. FLOPs are computed with an input image of size 1024 $\times$ 1024.}
  \centering
  \renewcommand\tabcolsep{2pt}
  \resizebox{\columnwidth}{!}{
  \begin{tabular}{c|c|c|c}
    \toprule
    Method  & Params(M) & FLOPs(T) & mDice $\uparrow$ \\
    \midrule
    SAM (ViT-H)~\cite{kirillov2023segment}& 635.6 & 2.73&0.778 \\
    EVP (ViT-H)~\cite{liu2023explicit}& 635.8& 2.74&0.849 \\
    SAM-adaptor (ViT-H)~\cite{chen2023sam}& 635.8& 2.74&0.850  \\
    \midrule
    MCA-SAM (ViT-B) (Ours)& 90.5 & 0.37 &0.836 \\
    MCA-SAM (ViT-L) (Ours)& 308.1 & 1.31&0.889  \\
    MCA-SAM (ViT-H) (Ours)& 635.8 & 2.74&\textbf{0.917} \\
    \bottomrule
  \end{tabular}
  }
  \label{tab:params}
\end{table}

\subsection{Parameters and FLOPs}
The comparison of parameters and FLOPs is shown in Table \ref{tab:params}. Based on the ViT-H backbone network, SAM~\cite{kirillov2023segment} achieves 0.778 mDice with 635.6M parameters and 2.73 TFLOPs.
Our MCA-SAM achieves 0.917 mDice with 635.8M parameters and 2.74 TFLOPs. The adaptors in MCA-SAM just include 0.2M parameters and the increasing FLOPs are 0.01 T.
These results demonstrate that our method is lightweight and efficient.

\begin{table}
  \caption{The ablation study of MCA-SAM (ViT-B) on Polyp~\cite{jha2020kvasir} dataset. SAM and SAM-adaptor are also based on the backbone network of ViT-B.}
  \centering
  \renewcommand\tabcolsep{5pt}
  \resizebox{\columnwidth}{!}{
  \begin{tabular}{c|c|c}
    \toprule
    Method  & mDice $\uparrow$ & mIoU $\uparrow$\\
    \midrule
    SAM  & 0.721 & 0.614 \\
    SAM + SC-adaptor & 0.819 & 0.694 \\
    SAM + TC-adaptor  & 0.829 & 0.709\\
    \midrule
    SAM + TC-adaptor + SC-adaptor (MCA-SAM)  & {0.836} & {0.748} \\
    \bottomrule
  \end{tabular}
  }
  \label{tab:ab_components}
\end{table}

\subsection{Ablation study}
\label{sec:exp_abalation}
TC-adaptor and SC-adaptor are the two main components of MCA-SAM. In the ablation study, we first verify the effectiveness of the two components. Then, we study the loss functions used in MCA-SAM and explore the different augmentation strategies used in SC-adaptors for contrastive learning. In the final, we compare the coverage speed of different adaptor contrastive learning strategies.

\subsubsection{The effectiveness of TC-adaptor and SC-adaptor}
Token-level Contrastive adaptor (TC-adaptor) and Sample-level Contrastive adaptor (SC-adaptor) are two main components of MCA-SAM, leading the cross-scale contrastive learning for SAM. The ablation study of the two components is shown in Table \ref{tab:ab_components}. The original SAM just reach mDice of 0.721 and mIoU of 0.614, which SAM-adaptor \cite{chen2023sam} achieves 0.807 in mDice and 0.678 in mIoU.
Then, MCA-SAM achieves mDice of 0.819 and mIoU of 0.694 by SC-adaptor, which verifies the effectiveness of token-level contrastive learning.
With the TC-adaptor, MCA-SAM achieves mDice of 0.829 and mIoU of 0.709, which demonstrates the significant improvements achieved by the TC-adaptor. 
Combined with SC-adaptor and TC-adaptor, MCA-SAM achieves SOTA results by the mDice of 0.838 and mIoU of 0.724.
These results demonstrate that integrating token-level and sample-level contrastive learning can significantly enhance the learning capability of the adaptor, thereby boosting the overall performance of SAM.

\begin{figure}[t]
  \centering
  \includegraphics[width=0.99\columnwidth]{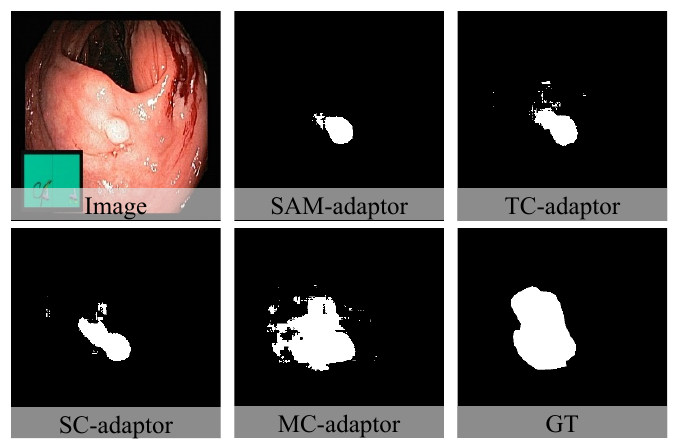}

  \caption{The visual comparison between SAM-adaptor, TC-adaptor, SC-adaptor, and MC-adaptor on the case from Polyp~\cite{jha2020kvasir} dataset. All models are based on the backbone network of ViT-B.}
  \label{fig:ab_vis}
\end{figure}

\begin{figure}[t]
  \centering
  \includegraphics[width=0.99\columnwidth]{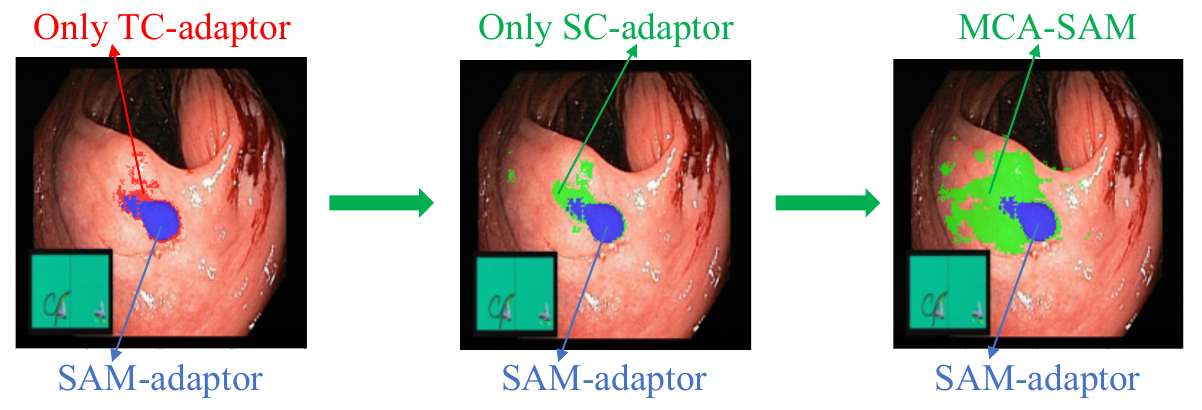}

  \caption{The TC-adaptor focuses on local details, while the SC-adaptor emphasizes more global context. MCA-SAM, which combines both the TC-adaptor and SC-adaptor, has a stronger ability to distinguish confusing areas and performs better than the SAM-adapter.}
  \label{fig:ab_vis2}
\end{figure}

The visual comparison between SAM-adaptor, TC-adaptor, SC-adaptor, and MC-adaptor (combining TC-adaptor and SC-adaptor) is shown in Figure \ref{fig:ab_vis}. Given the hard case like the image in Figure \ref{fig:ab_vis}, the SAM-adaptor just local the small part of the target area, while the MC-adaptor can recall more regions of interest since it has stronger discriminability brought by the proposed multi-scale contrastive adaptor learning. As shown in Figure \ref{fig:ab_vis2}, compared with the SAM-adaptor, MCA-SAM shows better results than the SAM-adapter.The TC-adaptor focuses on local details, while the SC-adaptor emphasizes more global context. MCA-SAM, which combines both the TC-adaptor and SC-adaptor, has a stronger ability to distinguish confusing areas and performs better than the SAM-adapter

\subsubsection{Loss Function}
In the experiments, except for the contrastive loss in the TC-adaptor and SC-adaptor, BCE loss is used to compute the error of predicted masks and ground-truth masks and IoU loss is used to achieve high values in evaluation metrics.
Table \ref{tab:ab_loss} shows the ablation study of used loss functions in MCA-SAM (ViT-B) on the Polyp dataset.
The MCA-SAM, employing both BCE loss and IoU loss concurrently, achieved the best performances in terms of mDICE and mIoU metrics. Consequently, we have adopted these two loss functions in all our experiments.

\begin{table}
  \caption{The ablation study of loss functions used in MCA-SAM (ViT-B) on Polyp~\cite{jha2020kvasir} dataset.}
  \centering
  \renewcommand\tabcolsep{13pt}
  \resizebox{\columnwidth}{!}{
  \begin{tabular}{c|c|c|c}
    \toprule
    Method  & Mask Loss & mDice $\uparrow$ & mIoU $\uparrow$\\
    \midrule
    MCA-SAM & BCE & 0.790 & 0.681 \\
    \midrule
    MCA-SAM & BCE + IoU & 0.836 & 0.748\\
    \bottomrule
  \end{tabular}
  }
  \label{tab:ab_loss}
\end{table}

 \begin{table}
  \caption{The ablation study of augmentations used in SC-adaptor of MCA-SAM (ViT-B) on Polyp~\cite{jha2020kvasir} dataset.}
  \centering
  \renewcommand\tabcolsep{12pt}
  \resizebox{\columnwidth}{!}{
  \begin{tabular}{c|c|c|c}
    \toprule
    Method  & Augmentation & mDice $\uparrow$ & mIoU $\uparrow$\\
        \midrule
    SAM & - & 0.721 & 0.614 \\
    SAM-adaptor & - & 0.807 & 0.678 \\
    \midrule
    \multirow{3}{*}{Single Aug.} & CJ & 0.775 & 0.678 \\
     & Gray & 0.800 & 0.706 \\
     & RS & 0.748 & 0.644 \\
    \midrule
    \multirow{4}{*}{Multiple Aug.} & CJ + RS & 0.836 & 0.748 \\
     & CJ + Gray & 0.774  & 0.674 \\
     & Gray + RS & 0.836 & 0.747  \\
     & CJ + RS + Gray & 0.769 & 0.669 \\
    \bottomrule
  \end{tabular}
  }
  \label{tab:ab_aug}
\end{table}

\begin{figure}[t]
  \centering
  \includegraphics[width=0.99\columnwidth]{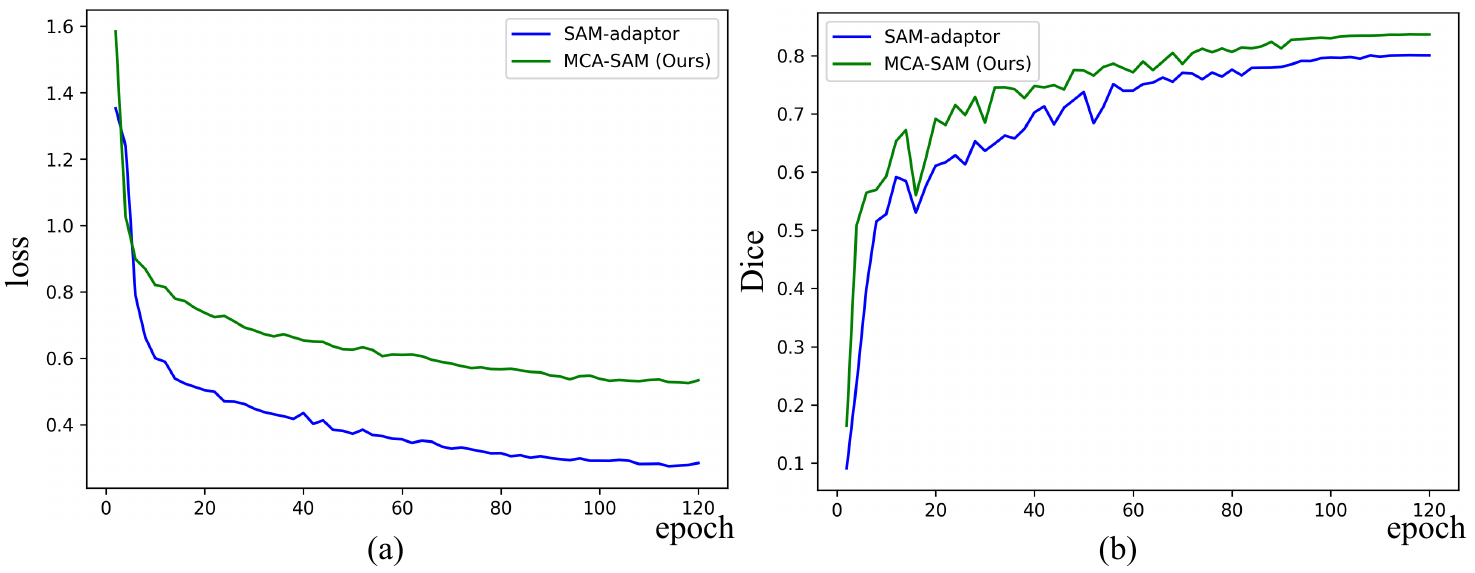}

  \caption{The convergence speed compared with SAM-adaptor~\cite{chen2023sam} based on backbone network of ViT-B.}
  \label{fig:ab_speed}
\end{figure}

\subsubsection{Different Augmentations for SC-adaptor}
The SC-adaptor is a crucial component within MCA-SAM that facilitates sample-level adaptor contrastive learning. It generates positive sample pairs by applying data augmentation to the original images, which are then utilized for computing the contrastive loss.
Table \ref{tab:ab_aug} demonstrates the ablation study of augmentations used in SC-adaptor. 
The used augmentations include ColorJitter (CJ), Grayscale (Gray), and Random Shfit (RS), and their combinations.
As shown in Table \ref{tab:ab_aug}, MCA-SAM fails to achieve satisfactory results if only a single augmentation strategy is employed in the SC-adaptor.
While the combination of multiple augmentation strategies can achieve better results. This phenomenon indicates that within the SC-adaptor's contrastive learning framework, the diversification of positive samples through generative methods can effectively bolster the performance of contrastive learning.
However, the different combinations of augmentations also impact the performance of MCA-SAM. For example, using the combination of ColorJitter and Grayscale, MCA-SAM just achieves 0.774 in mDice and 0.674 in mIoU, which is lower than other strategies. This is attributed to the fact that CJ and Gray exhibit similar characteristic variations. Thus, the combination of CJ, RS, and Gray also just achieves 0.769 in mDice and 0.669 in mIoU, respectively.

The main reason for these phenomena is that more diversified data augmentations can expand the contrast learning sample pool, so that adaptor can better learn the feature relationship between samples.
Meanwhile, the combination of CJ and Gray can't reach the best results since CJ and Gray are similar ways of augmentation that just change the color distribution of images, which limits the diversity of sample pairs.
In the experiments, the strategies of CJ + RS and Gray + RS can reach similar better results and our experiments are based on the strategy of CJ + RS. 

\begin{figure}[]
  \centering
  \includegraphics[width=0.99\columnwidth]{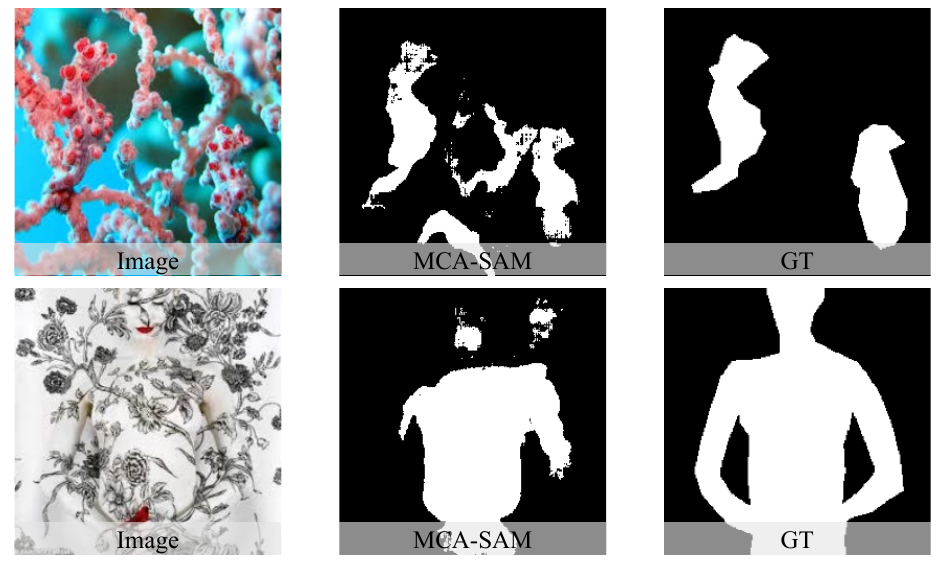}

  \caption{Some failure cases of MCA-SAM in the scene of extreme camouflage.}
  \label{fig:fail}
\end{figure}

\subsubsection{Convergence Speed}

The variation of loss and dice score during the training process is shown in Figure \ref{fig:ab_speed} (a) and Figure \ref{fig:ab_speed} (b), respectively.
MCA-SAM shows a bigger loss than SAM-adaptor in Figure \ref{fig:ab_speed} (a) since adding contrastive loss for adaptors. But MCA-SAM also shows a similar convergence speed with the SAM-adaptor.
As shown in Figure \ref{fig:ab_speed} (b), MCA-SAM shows better Dice scores during the training process.

\subsection{Failure Cases}
\label{sec:fail}
Although MCA-SAM can outperform previous adaptor-based methods, it still shows some failure in the scene of extreme camouflage.
As shown in Figure \ref{fig:fail}, it's difficult for MCA-SAM to distinguish these extremely confusing objects even for humans.
Meanwhile, it may caused by the limited data with annotations in these scenes. Thus, providing and annotating more high-quality data in these scenes may relieve this problem.

\section{Conclusion}
In this work, we investigate the adaptation of large-scale vision models to scenarios where performance is suboptimal. We introduce a novel and effective multi-scale contrastive adaptation framework for the Segmenting Anything Model (SAM), enhancing its transferability to challenging contexts. Our approach encompasses two distinct yet complementary strategies: token-level contrastive adaptation (TC-adaptor) and sample-level contrastive adaptation (SC-adaptor), which together aim to augment SAM's capacity and discriminative power. The synergy of TC-adaptor and SC-adaptor within our multi-scale contrastive adaptation framework facilitates efficient and robust contrastive representation learning. Comprehensive evaluations across four distinct datasets substantiate the superiority of our proposed method.











\bibliographystyle{cas-model2-names}

\bibliography{main}
\newpage

\end{document}